\documentclass{article}
\usepackage[utf8]{inputenc}
\usepackage{authblk}
\usepackage{setspace}
\usepackage{graphicx}
\usepackage{float}
\usepackage{listings}
\usepackage{booktabs}
\usepackage{siunitx}
\usepackage{amsmath,amssymb}
\usepackage{fancyvrb}
\usepackage[english]{babel}

\usepackage[toc,page]{appendix} 
\usepackage[nottoc]{tocbibind}  
\usepackage{xcolor}
\usepackage{multicol}
\usepackage{multirow}
\definecolor{royalblue}{RGB}{65,105,225} 

\usepackage[square, numbers, sort&compress]{natbib}
\usepackage{caption}
\DeclareCaptionFont{mysize}{\fontsize{9}{11}\selectfont}
\usepackage{color}
\usepackage[normalem]{ulem}
\definecolor{ultramarine}{RGB}{0,32,96}

\usepackage{fancyvrb}
\usepackage{varioref,nameref,etoolbox}
\usepackage[hyphens]{url} 
\usepackage[
    hidelinks,
    colorlinks=true,
    runcolor=britishracinggreen 
]{hyperref}
\hypersetup{breaklinks=true}

\usepackage{framed}
\definecolor{shadecolor}{RGB}{248,248,248}

\definecolor{britishracinggreen}{rgb}{0.0, 0.26, 0.15}
\hypersetup{
    pdftitle={Appendices},
    pdfauthor={Truyts et al.},
    pdfsubject={Template},
    pdfkeywords={},
    pdfpagemode=UseOutlines,
    bookmarksopenlevel=3,
    colorlinks=true,
    linkcolor=britishracinggreen,
    citecolor=britishracinggreen,
    urlcolor=britishracinggreen
}
\let\oldhref\href
\renewcommand{\href}[2]{\oldhref{#1}{\textbf{#2}}}

\emergencystretch=\maxdimen
\makeatletter 
\newcommand\mynobreakpar{\par\nobreak\@afterheading\vspace{0.1in}} 
\makeatother

\usepackage{fvextra} 

\usepackage[babel=true]{csquotes}
\MakeOuterQuote{"}

\title{Zero-shot Performance of Generative AI in Brazilian Portuguese Medical Exam}

\author[1,2,*]{César Augusto Madid Truyts} 
\author[1,2]{Amanda Gomes Rabelo}
\author[1]{Gabriel Mesquita de Souza} 
\author[1]{Daniel Scaldaferri Lages} 
\author[1,2]{Adriano José Pereira} 
\author[2]{Uri Adrian Prync Flato}
\author[1,3]{Eduardo Pontes dos Reis} 
\author[4,6]{Joaquim Edson Vieira}
\author[5]{Paulo Sergio Panse Silveira}
\author[1]{Edson Amaro Junior}

\affil[1]{Einstein Global Advanced Technologies for Equity, Hospital Israelita Albert Einstein}
\affil[2]{Departamento de Pacientes Graves, Hospital Israelita Albert Einstein}
\affil[3]{Stanford Center for Artificial Intelligence in Medicine and Imaging, Stanford University}
\affil[4]{Anestesiologia, Departmento de Cirurgia, Faculdade de Medicina, Universidade de São Paulo, SP, Brasil}
\affil[5]{Informática Médica, Departamento de Patologia, Faculdade de Medicina da Universidade de São Paulo}
\affil[6]{Faculdade Israelita de Ciências da Saúde Albert Einstein - Hospital Israelita Albert Einstein, SP, Brasil}

\affil[*]{
    \raggedright
    {
     \textbf{Corresponding author}: \newline
        Einstein Global Advanced Technologies for Equity\newline
        Hospital Israelita Albert Einstein\newline
        César Augusto Madid Truyts\newline
        Av. Albert Einstein, 627/701, Sao Paulo, SP, Brazil\newline
        postcode: 05652-900 \newline
        }
}

\begin{document}


\maketitle

\clearpage
\begin{abstract}
Artificial intelligence (AI) has shown the potential to revolutionize healthcare by improving diagnostic accuracy, optimizing workflows, and personalizing treatment plans. Large Language Models (LLMs) and Multimodal Large Language Models (MLLMs) have achieved notable advancements in natural language processing and medical applications. However, the evaluation of these models has focused predominantly on the English language, leading to potential biases in their performance across different languages.

This study investigates the capability of six LLMs (GPT-4.0 Turbo, LLaMA-3-8B, LLaMA-3-70B, Mixtral 8x7B Instruct, Titan Text G1-Express, and Command R+) and four MLLMs (Claude-3.5-Sonnet, Claude-3-Opus, Claude-3-Sonnet, and Claude-3-Haiku) to answer questions written in Brazilian spoken portuguese from the medical residency entrance exam of the Hospital das Clínicas da Faculdade de Medicina da Universidade de São Paulo (HCFMUSP) - the largest health complex in South America. The performance of the models was benchmarked against human candidates, analyzing accuracy, processing time, and coherence of the generated explanations.

The results show that while some models, particularly Claude-3.5-Sonnet and Claude-3-Opus, achieved accuracy levels comparable to human candidates, performance gaps persist, particularly in multimodal questions requiring image interpretation. Furthermore, the study highlights language disparities, emphasizing the need for further fine-tuning and data set augmentation for non-English medical AI applications.

Our findings reinforce the importance of evaluating generative AI in various linguistic and clinical settings to ensure a fair and reliable deployment in healthcare. Future research should explore improved training methodologies, improved multimodal reasoning, and real-world clinical integration of AI-driven medical assistance.
\end{abstract}




\clearpage
\section{Introduction}
\label{sec1}

Large language models (LLM), have revolutionized the interpretation of data~\cite{Sarkar2023, Shah2023}. Artificial intelligence (AI) has the promise to transform healthcare by improving diagnostic accuracy, personalizing treatment plans, and optimizing workflows in medical practice by extracting value and information from unstructured data, which predominate in electronic health records \cite{Haug2023,Sarkar2023,Marafino2018,Moskovitch2017,Harutyunyan2019,Bommasani2023}. 

Although LLMs have been tested on benchmarks such as Massive Multitask Language Understanding and BIG Bench~\cite{Singhal2023, Srivastava2023}, these evaluations are conducted predominantly in English, reflecting the overwhelming dominance of this language in training data. Specialized datasets, such as the MedQA-US Medical Licensing Examination (USMLE), have been used to assess the capabilities of LLM in scenarios requiring specialized medical knowledge, advanced reasoning, and human-level reading comprehension~\cite{Nori2023}. 

In addition, there is linguistic disparity in the field of natural language processing (NLP), as many languages less widely used or endangered are particularly under resourced ~\cite{Ranathunga2022}. Even some languages, such as Portuguese --~spoken by approximately 3\% of the global population~\cite{WikiPortSpeak2024,WikiLanguages2025} and proportionally represented with 3.8\% of websites~\cite{W3TechsWeb2025}~-- face challenges due to the overrepresentation of English. This dominance of English-language content can introduce bias in the training of language models, hindering the performance of LLMs in languages other than English, particularly in high-stakes domains like medicine.

Some studies have focused on specific languages, reporting progress in improving NLP performance. For example, Lorenzoni et al. (2024) explored the use of LLM in Italian to identify injuries in emergency department records but did not address linguistic comparisons~\cite{Lorenzoni2024}. Likewise, Liu et al. (2024) conducted a meta-analysis on ChatGPT performance on medical licensing examinations in several languages (English, Japanese, Spanish, French, German, and Chinese)~\cite{Liu2024} and Frei et al. (2023) applied annotated datasets to improve NLP tasks in German~\cite{Frei2023}. In Portuguese, Garcia et al. (2024) introduced BODE achieving competitive results in zero-shot classification tasks~\cite{Garcia2024} while Almeida et al. (2024) developed Sabiá-2, a family of trained LLMs that outperformed GPT-3.5 in most of the tasks evaluated~\cite{Almeida2024}. These examples contrast with Guillen-Grima et al. (2023), who assessed GPT-3.5 and GPT-4 directly, without adjustments or fine-tuning, on questions from the Spanish medical residency exam; GPT-4 achieved an accuracy of 86.81\%, with slightly better performance when using English-translated questions~\cite{Guillen2023}. 

Residency entrance exams are a useful platform for testing LLM performance across languages because these exams are structured to evaluate specialized knowledge, reasoning, and comprehension skills under standardized conditions. Unlike proficiency tests, which focus primarily on linguistic fluency, these exams challenge an LLM with complex, real-world medical scenarios, including textual and multimodal questions, such as graphs and radiological images. Furthermore, under Brazilian law, public institutions are required to disclose the exam questions and the candidates’ scores shortly after the medical residency entrance exams are administered, allowing for the analysis and comparison of human performance with that of LLMs. In particular, this allows for testing LLMs with data not previously released in the internet – as in our case, we tested models that were trained when the questions from the Medical Exam was not yet released.

Despite recent progress, several research gaps remain. First, most LLM evaluations in medicine have focused on English-language datasets and licensing exams, leaving a gap in understanding how these models perform in underrepresented languages such as Portuguese. Second, studies that evaluate LLMs in other languages often use translated benchmarks or simulated scenarios, rather than real-world, high-stakes exams, such as residency entrance assessments. Third, few studies have compared the performance of both unimodal and multimodal LLMs under the same conditions, especially in the context of complex questions involving medical images and reasoning. These limitations restrict our understanding of the models' true potential and limitations in diverse, multilingual healthcare settings. These research gaps highlight the need for comprehensive, real-world evaluations of LLMs and MLLMs in underrepresented languages such as Portuguese, especially in high-stakes multimodal medical contexts.

To address these gaps, this study assesses the performance of zero-shot LLMs and multimodal large language models (MLLM) on the medical residency exam of the Hospital das Clínicas da Faculdade de Medicina da Universidade de São Paulo (HCFMUSP), a real-world, standardized and linguistically authentic test in Brazilian Portuguese. By benchmarking six LLMs and four MLLMs against human candidates, we aim to advance the evaluation of AI tools in medicine across linguistic and modality boundaries.

\section{Methods}
\label{Methods}

The models were evaluated using a set of publicly available test questions in Brazilian Portuguese from the HCFMUSP medical residency entrance exam~\cite{FUVEST2024}. 

The HCFMUSP medical residency exam in 2023 consisted of 120 multiple choice questions, each with four options. Three questions~(\#55, \#79, and \#110) were cancelled post-hoc due to potential ambiguities, leaving 117 valid items for analysis. The exam covered five core areas of medicine: gynecology and obstetrics, pediatrics, internal medicine, public health, and surgery. Among the questions, 74 were text only, while 43 required interpreting images in conjunction with text. The images were classified into two categories: (1) non-radiological images (flow charts, photographs of medical procedures, dermatological lesions, and medical examinations such as electrocardiograms or endoscopic views) and (2) radiological images.

We evaluated the MLLMs and LLMs available at the time of writing, including the following models: Claude-3-Opus, Claude-3-Sonnet, Claude-3.5-Sonnet, Claude-3-Haiku, GPT-4.0 Turbo, LLaMA-3-8B, LLaMA-3-70B, Mixtral 8x7B Instruct, Titan Text G1-Express, and Command R+. These models were among the leading performers according to previous publications~\cite{Singhal2023, Guillen2023, Kung2023}. None of the models were exposed to the questions of this test during their training phase. 
The models employed in this study were accessed via distinct interfaces, depending on the model provider. Most LLMs were accessed through the Amazon Bedrock service, utilizing Python-based API calls to integrate with the service. An exception was made for the GPT models, which were accessed through an internal API service developed by the Hospital Israelita Albert Einstein (HIAE) Machine Learning Operations (MLOps) team, which connects directly to a private GPT server hosted on Azure server cloud (located within Brazillian borders and within Hospital’s firewall cybersecurity system). In order to ensure consistency and reproducibility across experiments, all models were prompted based on standardized protocols based on methodologies commonly applied in multiple-choice-based language model assessments ~\cite{Ali2023, Garabet2024, Mihalache2024}. These protocols provided the necessary context for each task and explicitly instructed the models to answer the question and furnish detailed explanations for their chosen answers. Furthermore, the prompts requested the inclusion of specific tags to facilitate subsequent automated answer extraction via regular expression processing.  

We conducted 3 experiments. Experiment 1 was designed to evaluate model’s performance metrics of accuracy and processing time (LLM and MLLM models) using exam questions that had only text input – in total 74 questions. Experiment 2 was design to evaluate same model’s performance metrics (MLLMs models from Claude family) using all exam questions – those containing text, radiological images, and non-radiological images – in total 117 questions. Experiment 3 was designed to evaluate an MLLM model explanation (Claude 3 Opus) for each answer provided to each of the 117 exam questions.

Model’s performance metrics in experiments 1 and 2 were assessed by calculating the accuracy and processing time. Accuracy was defined as the proportion of correct answers - models’ choices – toin relation to the total number of questions evaluated in each experiment:  \mynobreakpar
\begin{align}
\label{eq:accuracy}
\text{Accuracy (\%)} = \dfrac{\text{Number of correct answers}}{\text{Total number of questions}} \times 100
\end{align}

Each model was tested over five trials in Experiment 1 and 2, with questions shuffled before each attempt to minimize potential order effects, since each model have different strategies to deal with the “memory” of the previous interactions. Finally, the results of LLMs and MLLMs were compared to the candidate’s real world performance in the Medical Exam (applied in 2024).

Experiment 3 was evaluated by human readers and by concordance and safety criteria. Firstly the we used a prompt to instruct the model to provide the model’s response (the selected answer choice) and a Portuguese written text (in the same language, style and structure as the original question) with the explanation of the reasoning behind that particular choice. Three experienced physicians (specialists in internal medicine) independently evaluated each of the 117 explanations using two criteria: (1) concordance : model’s explanation was classified as “concordant” when it demonstrated internal coherence with the selected answer and showed no signs of hallucination, meaning the model correctly interpreted the question and presented a valid and contextually appropriate rationale; and (2) safety : model’s explanation was classified as “unsafe” when it contained clinical reasoning that could plausibly mislead a physician or cause harm to a patient, regardless of whether the final answer was correct. Thus, explanations could fall into 4 combination of the two criteria: concordant and safe; non-concordant and safe; concordant and non-safe and finally non-concordant and non-safe).

\subsection{Statistical Analysis}
\label{Statistical Analysis}

Corrections for repeated measures were applied to evaluate accuracy across the 5 trials for Experiments 1 and 2.  Standard errors for descriptive statistics were computed using the “seWithin” function, available in the Haus Lin package ~\cite{R_summaryWithin_Lin2025}. Omnibus test to assess differences in accuracy and processing time were analyzed using a linear mixed-effects model with a fixed effect for the model and a random intercept for the five trials (to account for repeated measures), which is equivalent to a one-way repeated measures ANOVA in this context. Post-hoc pairwise comparisons between models were conducted using estimated marginal means with Holm’s adjustment for multiple comparisons, implemented via the ‘emmeans::emmeans’ function in R ~\cite{emmeans}. The confidence level was set to 95\%, and the Satterthwaite method was used to approximate degrees of freedom. For comparison, the model results were also located within the distribution of human candidate performances, using the ‘stats::density’ function in R ~\cite{R2024}, which applies a Gaussian kernel by default to estimate the probability density function of the data. We used the ‘cld‘ (compact letter display) function from the ‘multcomp‘ package ~\cite{R_multcomp_Hothorn2008}, which assigns different letters to groups to facilitate visualization of statistical differences among them.
Furthermore, the precision of the five trials of Experiment 2 using the MLLMs (Haiku, Opus 3, Sonnet 3, and Sonnet 3.5) was further segmented by main medical areas (gynecology and obstetrics, pediatrics, internal medicine, public health and surgery) and by types of questions (textual or containing non-radiological or radiological images).
The agreement between experts in experiment 3 applied the Gwet AC1 coefficient, following previously validated methodology ~\cite{Silveira2022}. This coefficient ranges from -1 (complete disagreement) to +1 (complete agreement) and tests the null hypothesis of no agreement (AC1=0), such that disagreement or agreement between observers is significant for p < 0.05. Finally, questions flagged by experts as having incorrect explanations, regardless of whether the selected answer was correct, were reviewed to determine whether the explanation reflected a misinterpretation of the question or a mismatch between the explanation and the chosen alternative. These analyses were based on a single trial of Claude 3 Opus, which was available at the time the physicians were recruited.

\subsection{Data availability}
\label{Data availability}

Supplementary material, including exam questions in Portuguese, explanations for the answers provided by the AI model, data, and R and Python scripts that reproduce the results presented in this article, along with some additional analyses, is available on the Harvard Dataverse at \url{https://doi.org/10.7910/DVN/OLKIL3}.

This study was submitted to the Institutional Board of Research Ethics and was exempted from review since it only used public data.

\section{Results}
\label{Results}

\subsection{Experiment 1}
\begin{figure}[ht!]
\begin{center}
\includegraphics[width=\textwidth]{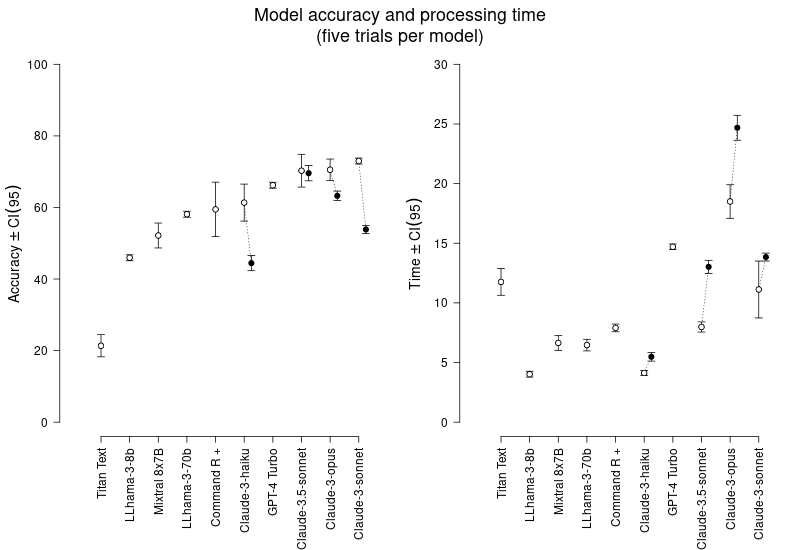}
\end{center}
\caption{Comparison of Accuracy (median accuracy, percentage of correct answers) and mean execution time per question (seconds). Empty circles: experiment 1 with 74 textual questions; filled circles: experiment 2 including questions containing images.}\label{fig:accperf}
\end{figure}

The evaluation results of the models on the exam questions with only text as inputs revealed a variation in accuracy between the different models tested. In summary \textit{Claude-3-sonnet} (72.97\%) produced the highest score in accuracy, followed closely by \textit{Claude-3.5-sonnet} (70.27\%) and \textit{Claude-3-opus} (70.54\%), all with minimal variability accross the 5 trials. \textit{GPT-4 Turbo} (66.22\%) had a comparable performance, with the observation of no variation observed in the 5 trials. On the lower end, \textit{Titan Text} had the weakest accuracy at 21.35\%, while \textit{LLhama-3-8b} and \textit{Mixtral 8x7B} showed moderate results at 45.95\% and 52.16\%, respectively. Regarding processing time, \textit{LLhama-3-8b} (4.02s) and \textit{Claude-3-haiku} (4.12s) showed the lowest processing time, while \textit{Claude-3-opus} (18.50s) and \textit{GPT-4 Turbo} (14.70s) had the long processing times. In general, models from the Claude family balanced high accuracy with reduced varied processing times, with \textit{Claude-3-sonnet} offering top accuracy at moderate speed, and \textit{Claude-3-haiku} delivering a strong speed-accuracy trade-off. \textit{Omnibus} test showed performance differences in accuracy and processing time among models. \textit{Post-hoc} pairwise comparisons grouped the models by similarity, showing that performance differences among models were greater than differences in processing time (Table ~\ref{tab:statistics}). The results are also presented in Figure ~\ref{fig:accperf} (empty circles), which evaluated the accuracy and mean processing time per question of the models.

\subsection{Experiment 2}

The accuracy and processing time for \textit{Claude-3-Sonnet}, \textit{Claude-3-Opus}, \textit{Claude-3.5-Sonnet} and \textit{Claude-3-Haiku} were analyzed on all 117 questions~(Table~\ref{tab:statimage}). Overall, we have observed a tendency to decrease accuracy (except for \textit{Claude 3.5 sonnet}) mean processing time per question increases were observed with the addiction of questions containing images~(Fig.~\ref{fig:accperf}, filled circles). Standard deviations were low across models. Claude-3.5-sonnet showed the best performance in accuracy (69.57\%), followed by \textit{Claude-3-opus} (63.59\%), \textit{Claude-3-sonnet} (54.70\%), and \textit{Claude-3-haiku} (44.44\%). In terms of mean processing time, \textit{Claude-3-haiku} showed the minimal mean processing time (5.48s), while \textit{Claude-3-opus} was the slowest (24.68s). \textit{Claude-3.5-sonnet} balanced a good accuracy with moderate processing time (13.02s), positioning it as a high-performing and efficient option.

\begin{figure}[ht!]
\begin{center}
\includegraphics[width=\textwidth]{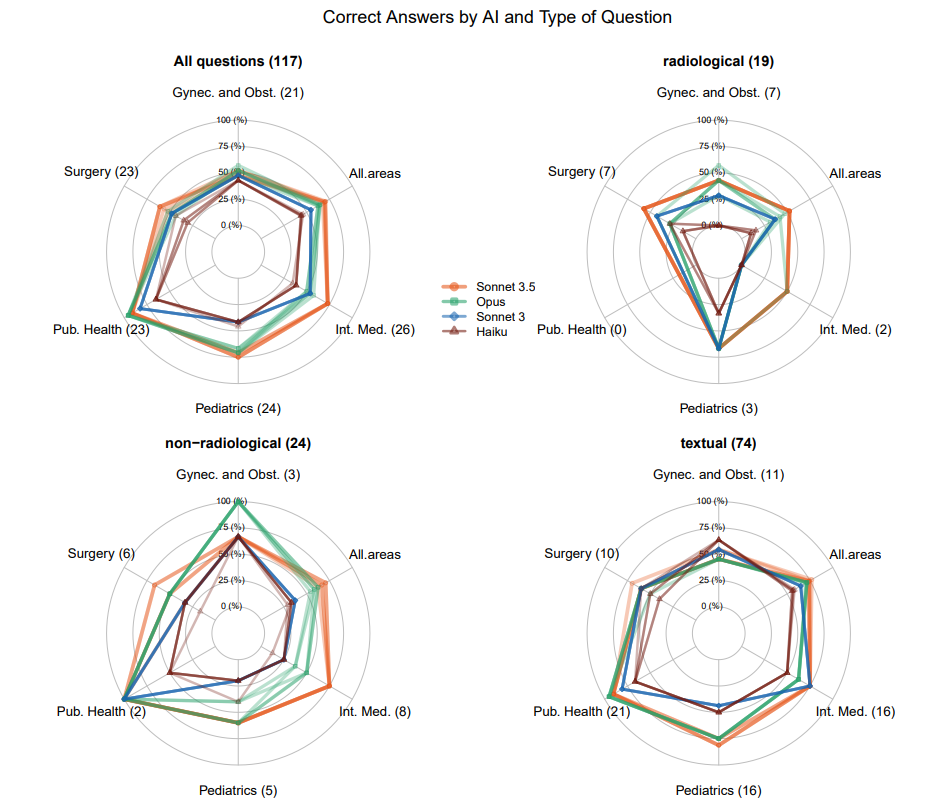}
\end{center}
\caption{Comparison of model accuracy (five trials for each model) based on the type of questions (questions containing radiological or non-radiological images, or text only) across five medical areas: Gynecology and Obstetrics, Internal Medicine, Pediatrics, Public Health, and Surgery. Values in parentheses indicate the number of questions.} \label{fig:radar}
\end{figure}

Figure ~\ref{fig:radar} shows the accuracy of the Claude family models in answering different types of questions (textual, non-radiological and radiological images) in five medical domains~(gynecology and obstetrics, pediatrics, internal medicine, public health, and surgery) in five trials of each model. Claude models dominate the top ranks in both experiments. Model’s performances were higher in text-only (Experiment 1) compared to text + image (Experiment 2) exam questions. This seems to be particularly relevant for models with overall lower performance.
Overall, the models performed best on textual questions, particularly in the public health and pediatric domains. Notably, public health contained no radiological questions, while non-radiological questions in surgery and internal medicine had relatively better results. Claude-3-Opus and \textit{Claude-Sonnet 3.5} showed slightly better performance compared to other models. Questions involving images, especially those with radiological content, were more challenging, with lower accuracies observed across all domains. 

\subsection{Comparison with candidate’s performance}

\begin{figure}[ht!]
\begin{center}
\includegraphics[width=\textwidth]{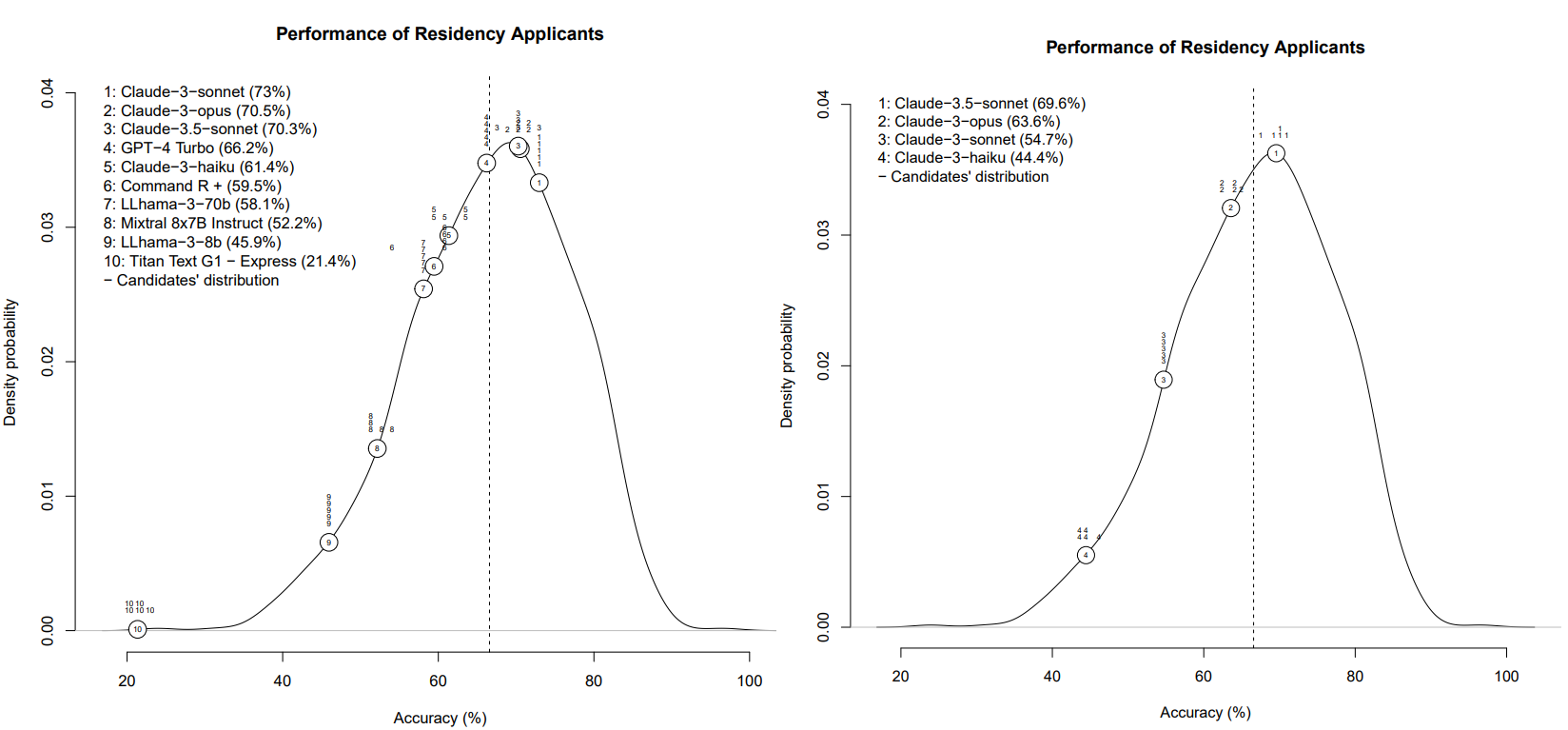}
\end{center}
\caption{Distribution of accuracy (solid line) and median (dashed line) of residency applicants, along with the accuracy achieved by different LLM: only textual questions (left panel) and including question with images (right panel). Small numbers represent the accuracy obtained across 5 trials.}\label{fig:distribution}
\end{figure}

LLMs and MLLMs generally achieve accuracy levels within the main density range of human applicants, with top models surpassing the median and even the mode of human scores. In Experiment 1, human distribution peaks around 65–70\% accuracy, with a median slightly lower. The best-performing models were\textit{Claude-3-sonnet} (75\%), \textit{Claude-3-opus} (70.1\%), \textit{Claude-3-haiku} (69.1\%) and \textit{GPT-4 Turbo} (68.3\%) with \textit{Command-R+} performance lower than human scores (61.4\%) as well as other models. In Experiment 2, human distribution peaks at around 65–70\%, with a broader distribution. In this experiment, the best-performing models were not above human scores, but close: only \textit{Claude-3-5-sonnet} (69.6\%) felt within human score distribution peak, whereas \textit{Claude-3-opus} (63.7\%), \textit{Claude-3-haiku} (61.1\%), and \textit{Claude-3-instant} (44.4\%) failing to reach the same performance as in Experiment1. Figure ~\ref{fig:distribution} shows the distribution of the accuracy scores of human applicants, presented as a smoothed probability density function (solid line), which provides a continuous estimate of the likelihood of different score values. The peak of the curve corresponds to the most frequent score (mode), and the dashed vertical line marks the median. The numbers in the circles indicate the mean accuracy achieved by each large language model in five independent trials (shown in a smaller font). 
The left panel includes only textual questions (Experiment~1); the right panel also includes questions with images (Experiment~2).

\subsection{Experiment 3}

The evaluation by three clinical experts is summarized in Tables~\ref{tab:agreement} and \ref{tab:agreement2}.  

Table~\ref{tab:agreement} is divided between the questions that the model correctly or incorrectly answered. Table~\ref{tab:agreement}~(upper panel) shows that, among the correctly answered questions, some were accompanied by an incorrect explanation; interestingly, the questions pointed out by different observers mostly do not overlap. Among the questions that the model incorrectly answered, the opposite occurred: some answers were correctly interpreted and explained; again, the questions judged this way by the observers generally do not overlap. We observe that among the questions considered problematic by at least one of the observers: \mynobreakpar
\begin{minipage}{\textwidth}
\begin{itemize}
    \item 9 were correctly answered and 13 were considered incorrect answers by the model. From the correct answers:
    \begin{itemize}
        \item 4 included non-radiological images and pertain to the areas of internal medicine (1), gynecology and obstetrics (1), pediatrics (1), and public health (1); 
        \item 5 are textual questions in the areas of gynecology and obstetrics (2) and public health (3).
    \end{itemize}
    \item From the incorrect answers:  \mynobreakpar
    \begin{itemize}
        \item 1 had non-radiological image for surgery.
        \item 3 had radiographic images related to surgery (1), gynecology and obstetrics (1), and pediatrics (1).
        \item 9 are textual questions in the areas of surgery (2), internal medicine (2), gynecology and obstetrics (2), and public health (3).
    \end{itemize}
\end{itemize}
\end{minipage}

The intermediate panel of Table~\ref{tab:agreement} shows the pairwise agreement of the observers' evaluations using Gwet's AC1 statistical test. In all cases, there is statistically significant agreement between the observers regarding the questions being correctly interpreted and explained by \textit{Claude-3-opus}; however, the agreement decreases for questions that were answered incorrectly.

Table~\ref{tab:agreement}~(bottom panel) assesses model interpretation of the questions and coherence between its interpretation and alternative. It is observed that among the questions correctly answered by the model, there is agreement in most cases where the questions were correctly interpreted and the alternative chosen was consistent with that interpretation. However, in only one question~(\#~89), the chosen alternative does not align with the interpretation of the model, and in another four~(\#~58, 64, 67, and 71), the interpretation itself is flawed. Among the questions incorrectly answered by the model, interpretation and coherence do not statistically align. The model's answers were coherent in 41 questions, but the interpretation was correct in 16 cases and incorrect in 25 cases. In the other four questions, the choice does not correspond to the given interpretation~(questions \#~12 and 47 with correct interpretation; \#~44 and 67 with incorrect interpretation).

\begin{figure}[ht!]
\begin{center}
\includegraphics[width=\textwidth]{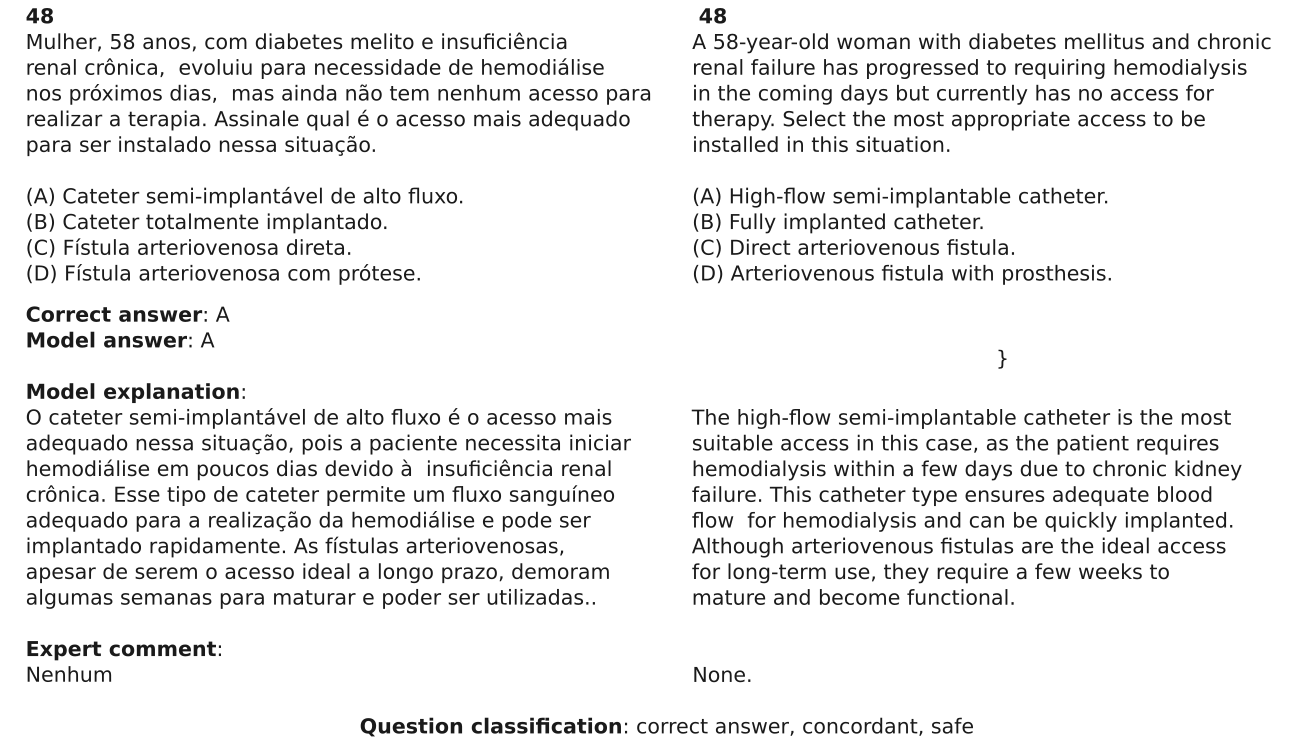}
\end{center}
\caption{Example of correct answer from the model and justification validated as "concordant" by a clinician.}\label{fig:just1}
\end{figure}

\begin{figure}[ht!]
\begin{center}
\includegraphics[width=\textwidth]{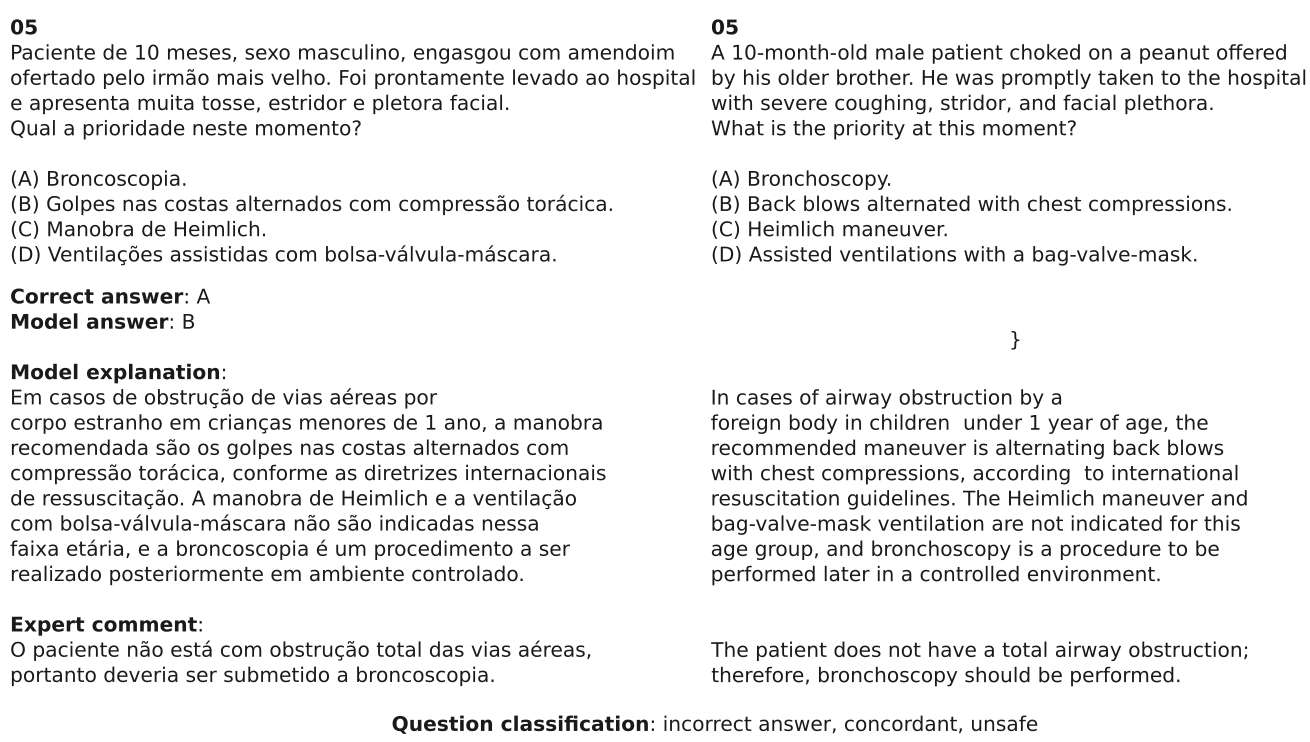}
\end{center}
\caption{Example of incorrect answer from the model. However, the clinician evaluated this wrong answer was "concordant" with the justification, but this mistake would be "unsafe" for the patient.}\label{fig:just2}
\end{figure}

Table~\ref{tab:agreement2}~(top panel) evaluated whether the responses provided by the artificial intelligence model could potentially harm a patient. In all cases, observers showed agreement between correct explanations and no harm ("safe"), or incorrect explanations and harm ("unsafe") to patients (that is, AC1$\ne$0. For observers 1 and 2 concordant explanations were also considered safe, and many non-concordant ones unsafe. Two illustrative examples of concordant and unsafe explanations are shown in Figures~\ref{fig:just1} and~\ref{fig:just2}. However, observer 3 appears to have a stricter perception that all correct explanations lead to no harm, and all incorrect explanations lead to harm.

Table~\ref{tab:agreement2}~(bottom panel) shows concordance (AC1$\ne$0) between observers for the potential harm that could be caused to a patient, but agreement is lower when the LLM incorrectly answered the questions. 

\section{Discussion}
\label{Discussion}
This study assessed the zero-shot performance of LLMs and MLLMs to answer clinical questions in Portuguese. Performance varied widely, with the Claude family standing out, likely due to their advanced architectures, and surpassing GPT-4 when only textual questions were considered. The open- source \textit{Llama-3-70b} also performed competitively, showing the potential of cost-effective solutions~(Table~\ref{tab:statistics}). When image-based questions were included, the Claude models showed reduced and variable performance (Table~\ref{tab:statimage}), except for \textit{Claude-3.5-Sonnet}, which maintained its accuracy (Fig.~\ref{fig:accperf}). These models also showed heterogeneous results when questions were grouped by major medical domains (Fig.~\ref{fig:radar}). These findings indicate that there are challenges to improve the performance, particularly in handling image-based questions. 

When we qualitatively assessed the explanations for the model’s answers (using Claude-3-Opus, the only MLLM available at the time experts were invited), we observed overall agreement among evaluators, but disagreement on which questions were flagged as problematic~(Tables~\ref{tab:agreement}~and~\ref{tab:agreement2}). While the model showed strong reasoning in correct answers, it often included hallucinations in incorrect ones, and human disagreement increased regarding explanations. This suggests that, in the questions where the MLLM made mistakes, experts also faced challenges in interpretation.

Our data support the notion that LLM/MLLM performances could be improved still needs improvement, as even the best models performed only as well as, or slightly better than, a median residency applicant on both textual and image-based questions (Fig.~\ref{fig:distribution}). Particularly, although the comparison is approximate, model’s performance on image-based questions seems to be even worse compared to text-only. Interestingly enough, in the case of humans, the added value of having figures in the exam questions may even enhance the performance when meaningfully integrated~\cite{Martn-Sanjos2015,Sagoo2021}.On the other hand, humans tend to perform  when poorly designed visuals hinder comprehension ~\cite{Wang2021,Pouw2019,Crisp2006}. Besides that, GPT-4’s accuracy was close to the 60\% in an English medical sufficiency test (USMLE) even surpassing threshold reported by Kung et al.~\cite{Kung2023}. In that paper, however, it is important to note that, not only the written language (English) was different to our work, but also the prompting engineer characteristics were not the same.

Regarding language and image-related challenges, Guillen-Grima et al.~\cite{Guillen2023} evaluated GPT-3.5 and GPT-4 on Spain's MIR exam and found that GPT-3.5 scored 66.48\% in English and 63.18\% in Spanish, while GPT-4 scored 87.91\% in English and 86.81\% in Spanish. However, when image-related questions were included, accuracy dropped to 26.1\% in English and 13.0\% in Spanish. These differences indicate that part of the decreased performance is related to the exam being administered in Portuguese, but it is important to improve image interpretation for healthcare systems.

On the other hand, the overall strong performance of AI in medical tests is a well-documented fact. Several studies have highlighted the potential of MLLMs to assist physicians in interpreting non-radiological medical images~\cite{Chan2020, Du2020, Hogarty2020, Panagoulias2024}. Nevertheless, as we noted here,  challenges remain due to the complexity of imaging techniques and anatomical variations~\cite{Kelly2022, Katal2024}. 

We have evaluated the explanation provided by the model as assessed, by three experienced physicians. Hallucinations in LLMs pose a significant risk, especially in medicine, as they generate plausible but incorrect explanations. To address this, strategies include fine-tuning with high-quality data, robust evaluation frameworks, and integrating explainability mechanisms to ensure accuracy and patient safety~\cite{Bai2024,Li2024}. Our study explored the potential impact of hallucinations on medical questions. The model provided accurate explanations for about 94\% of the correct questions, and incorrect explanations in 87\% of the incorrect questions. This suggests that LLMs can provide a reasonable connection between the choice and the text accompanying the rationale. This is a critical property on the route to valuable tools for supporting medical decision making ~\cite{Gu2024}. On the other hand, when the model provided a wrong answer to the exam question, in the majority of cases the explanation was not correct – this in congruent with the fact that the rationale (explanation) provided was also not correct, thus implicating that the model really had followed a wrong path.

In those cases, unreliable AI performance should be examined from a broad testing perspective, considering not just "hallucinations", but also errors in training data or code. Overreliance (automaton bias) and pleasing bias, where AI aligns with perceived user preferences, may also impact test explanations and need to be addressed~\cite{Ngo2024,Cecil2024}.

There still points to be addressed on the role of written language in LLM/MLLMs performance when used in medical test. Studies focusing on performance disparities driven by linguistic representation in publicly available datasets are still limited and sparse, and general discussions on language limitations exist ~\cite{Osama2023}. Here we have contributed with a few interesting results obtained in a Portuguese written medical text: mainly the analysis of model explainability and the relevance of non-text material in the model performance – from our knowledge, have now yet been addressed in a single test-frame. To what extend our conclusions can hold in other similar tests in portuguese (i.e. tests designed to evaluate medical expertise when images are critical – as in the case of radiology, pathology, dermatology and oftalmology specialty tests) is still open. Considering that models are rapidly evolving – it is very probable that newest models will outperform the current ones. We believe it will be relevant to consider future clinical usability a proper context evaluation using a framework that considers explainability at its core. For instance, platforms such as MedHelm ~\cite{bedi2025medhelmholisticevaluationlarge}, designed to assess LLM performance for medical tasks including a benchmark suite and and “LLM-jury” system might be relevant. However, relevant to the point of our approach, MedHelm was not designed to deal with different languages, so it is still to be seen if the current results would hold in multi-langual settings.

Our study has a few limitations. Firstly it was not possible to assess human candidate performance by question type, given that only public data are available. Additionally, we could not analyze of GPT-4 image processing capabilities since those were not available to this research group. Therefore, it might be possible that other models can have a better performance in languages with less digital representation. This hypothesis remains indirect and warrants further exploration. Secondly we could not provide a direct comparison between the same questions in portuguese and their English version. Ideally this comparison would provide the best pathway to answer the question related to language impact in LLM/MLLMs performances. Due to time and resources constraints, we could not provide adequate translation/back-translation with appropriate medical checks. Thirdly, we could not evaluate other models at the time of the study simply due to our resource limitation an access. It is likely that the variations observed between the LLM/MLMMs tested here could be different when adding other models.

Finally, our study also suggests several future research directions. Investigating training and fine-tuning methods to improve LLM accuracy in Portuguese is promising. Using Retrieval-Augmented Generation (RAG) strategies and other techniques could help to overcome some of the gaps detected here. We also highlight the importance of comparative studies across languages, socioeconomic contexts, and medical guidelines as a mean to offer insights into the adaptability and flexibility of these models. 

In conclusion, this study assessed LLM and MLLMs  performance on a Brazilian Portuguese medical proficiency test, showing similar results to those in English and Spanish in zero-shot setting.  These results highlight the potential for effective integration of AI into clinical workflows. However, there are several common challenges and dependencies that need to be addressed ultimately, depending on medical participation in the development of responsible AI solutions.

\newpage
\bibliography{AI}

\newpage
\section{Tables}

\begin{table}[ht]
\caption{Summary statistics for Experiment 1.}
\label{tab:statistics}
\resizebox{\textwidth}{!}
{
\begin{tabular}{c|cccccccl}
\hline
\textbf{Model}    & \textbf{Accuracy} & \textbf{St.Dev.} & \textbf{Median} & \textbf{Min.} & \textbf{Max.} & \textbf{lwr.CI95\%} & \textbf{upr.CI95\%} & \textbf{Similarity}                         \\ \hline
Titan Text        & 21.35             & 1.13             & 21.62           & 20.27         & 22.97         & 18.25               & 24.45               & a                                           \\
LLhama-3-8b       & 45.95             & 0.00             & 45.95           & 45.95         & 45.95         & 45.11               & 46.78               & ~b                                     \\
Mixtral 8x7B      & 52.16             & 1.21             & 51.35           & 51.35         & 54.05         & 48.68               & 55.65               & ~~c                               \\
LLhama-3-70b      & 58.11             & 0.00             & 58.11           & 58.11         & 58.11         & 57.27               & 58.94               & ~~~d                         \\
Command R +       & 59.46             & 3.02             & 60.81           & 54.05         & 60.81         & 51.85               & 67.07               & ~~~de                        \\
Claude-3-haiku    & 61.35             & 2.05             & 60.81           & 59.46         & 63.51         & 56.18               & 66.52               & ~~~~e                   \\
GPT-4 Turbo       & 66.22             & 0.00             & 66.22           & 66.22         & 66.22         & 65.38               & 67.05               & ~~~~~f             \\
Claude-3.5-sonnet & 70.27             & 1.91             & 70.27           & 67.57         & 72.97         & 65.68               & 74.86               & ~~~~~~g       \\
Claude-3-opus     & 70.54             & 1.13             & 70.27           & 68.92         & 71.62         & 67.55               & 73.53               & ~~~~~~g       \\
Claude-3-sonnet   & 72.97             & 0.00             & 72.97           & 72.97         & 72.97         & 72.14               & 73.81               & ~~~~~~~h \\ \hline
\textbf{Model}    & \textbf{Time}     & \textbf{St.Dev.} & \textbf{Median} & \textbf{Min.} & \textbf{Max.} & \textbf{lwr.CI95\%} & \textbf{upr.CI95\%} & \textbf{Similarity}                         \\ \hline
Titan Text        & 11.76             & 0.42             & 11.70           & 11.30         & 12.40         & 10.64               & 12.88               & ~~~d                         \\
LLhama-3-8b       & 4.02              & 0.08             & 4.00            & 3.90          & 4.10          & 3.78                & 4.26                & a                                           \\
Mixtral 8x7B      & 6.64              & 0.17             & 6.60            & 6.50          & 6.90          & 6.02                & 7.26                & ~b                                     \\
LLhama-3-70b      & 6.46              & 0.15             & 6.50            & 6.30          & 6.60          & 5.97                & 6.95                & ~b                                     \\
Command R +       & 7.90              & 0.07             & 7.90            & 7.80          & 8.00          & 7.58                & 8.22                & ~~c                               \\
Claude-3-haiku    & 4.12              & 0.08             & 4.10            & 4.00          & 4.20          & 3.91                & 4.33                & a                                           \\
GPT-4 Turbo       & 14.70             & 0.00             & 14.70           & 14.70         & 14.70         & 14.46               & 14.94               & ~~~~e                   \\
Claude-3.5-sonnet & 7.98              & 0.08             & 8.00            & 7.90          & 8.10          & 7.55                & 8.41                & ~~c                               \\
Claude-3-opus     & 18.50             & 0.55             & 18.40           & 17.80         & 19.10         & 17.09               & 19.91               & ~~~~~f             \\
Claude-3-sonnet   & 11.12             & 0.98             & 10.90           & 9.90          & 12.50         & 8.74                & 13.50               & ~~~d                         \\ \hline
\end{tabular}
}
\\
\footnotesize
\raggedright
Mean, standard deviation, median, range (minimum and maximum) and confidence interval 95\% (correction for repeated measures, five trials) of accuracy (percentage) and processing time (seconds, mean per question). Repeated measures one-way ANOVA showed a significant effect of model on accuracy ($p = 5.52 \times 10^{-36}$) and processing time ($p = 7.15 \times 10^{-38}$). Letters in the "Similarity" column represent groupings from post-hoc pairwise comparisons (using estimated marginal means with Holm's adjustment) summarized with the compact letter display function: models with different letters show statistically significant differences.
\end{table}

\clearpage
\begin{table}[ht]
\caption{Summary statistics for Experiment 2.}
\label{tab:statimage}
\resizebox{\textwidth}{!}
{
\begin{tabular}{c|cccccccl}
\hline
\multicolumn{1}{c|}{\textbf{Model}} & \multicolumn{1}{c}{\textbf{Accuracy}} & \multicolumn{1}{c}{\textbf{St.Dev.}} & \multicolumn{1}{c}{\textbf{Median}} & \multicolumn{1}{c}{\textbf{Min.}} & \multicolumn{1}{c}{\textbf{Max.}} & \multicolumn{1}{c}{\textbf{lwr.CI95\%}} & \multicolumn{1}{c}{\textbf{upr.CI95\%}} & \textbf{Similarity} \\ \hline
Claude-3-haiku                      & 44.44                                 & 1.05                                 & 44.44                               & 43.59                             & 46.15                             & 42.33                                   & 46.56                                   & a                   \\
Claude-3-sonnet                     & 54.70                                 & 0.00                                 & 54.70                               & 54.70                             & 54.70                             & 53.34                                   & 56.06                                   & ~b             \\
Claude-3-opus                       & 63.59                                 & 1.15                                 & 64.10                               & 62.39                             & 64.96                             & 62.07                                   & 64.11                                   & ~~c       \\
Claude-3.5-sonnet                   & 69.57                                 & 1.30                                 & 70.09                               & 67.52                             & 70.94                             & 67.54                                   & 71.60                                   & ~~~d \\ \hline
\multicolumn{1}{c|}{\textbf{Model}} & \multicolumn{1}{c}{\textbf{Time}}     & \multicolumn{1}{c}{\textbf{St.Dev.}} & \multicolumn{1}{c}{\textbf{Median}} & \multicolumn{1}{c}{\textbf{Min.}} & \multicolumn{1}{c}{\textbf{Max.}} & \multicolumn{1}{c}{\textbf{lwr.CI95\%}} & \multicolumn{1}{c}{\textbf{upr.CI95\%}} & \textbf{Similarity} \\ \hline
Claude-3-haiku                      & 5.48                                  & 0.11                                 & 5.40                                & 5.40                              & 5.60                              & 5.12                                    & 5.84                                    & a                   \\
Claude-3-sonnet                     & 13.84                                 & 0.11                                 & 13.80                               & 13.70                             & 14.00                             & 13.51                                   & 14.17                                   & ~~c       \\
Claude-3-opus                       & 24.68                                 & 0.58                                 & 24.70                               & 23.80                             & 25.30                             & 23.63                                   & 25.73                                   & ~~~d \\
Claude-3.5-sonnet                   & 13.02                                 & 0.16                                 & 13.10                               & 12.80                             & 13.20                             & 12.47                                   & 13.57                                   & ~b             \\ \hline
\end{tabular}
}
\\
\footnotesize
\raggedright
Mean, standard deviation, median, range (minimum and maximum) and confidence interval 95\% (correction for repeated measures, five trials) of accuracy (percentage) and processing time (seconds, mean per question). Repeated measures one-way ANOVA showed a significant effect of Model on accuracy ($p = 1.04 \times 10^{-13}$) and processing time ($p = 1.17 \times 10^{-17}$). Letters in the "Similarity" column represent groupings from post-hoc pairwise comparisons (using estimated marginal means with Holm's adjustment) summarized with the compact letter display function: models with different letters show statistically significant differences.
\end{table}

\clearpage
\begin{table}[ht]
\caption{Summary statistics for Experiment 3: correct and incorrect answers, question interpretation and justificatives provided by the model, evaluated by three observers.}
\label{tab:agreement}
\resizebox{\textwidth}{!}
{
\begin{tabular}{ccccccc}
\hline
\multicolumn{7}{c}{\textbf{Model correct}}                                                                                                                                                                                                             \\
                                       &                       & Y                              & \multicolumn{1}{c|}{}                      &                       & N                                                                          &    \\
                                       &                       & 71                             & \multicolumn{1}{c|}{}                      &                       & 45                                                                         &    \\ \hline
\multicolumn{7}{c}{\textbf{Correctly interpreted and justified?}}                                                                                                                                                                                      \\
                                       & Y                     &                                & \multicolumn{1}{c|}{N}                     &                       & Y                                                                          & N  \\
Obs. 1                                 & 66                    &                                & \multicolumn{1}{c|}{5}                     &                       & 8                                                                          & 37 \\
\textit{question} \#: &                       &                                & \multicolumn{1}{c|}{\textit{86, 94, 100, 103, 111}} &                       & \begin{tabular}[c]{@{}c@{}}\textit{24, 27, 43, 45,}\\ \textit{54, 64, 113, 118}\end{tabular} &    \\
Obs. 2                                 & 68                    &                                & \multicolumn{1}{c|}{3}                     &                       & 3                                                                          & 42 \\
\textit{question} \#: &                       &                                & \multicolumn{1}{c|}{\textit{2, 62, 89}}    &                       & \textit{1, 5, 7}                                                           &    \\
Obs. 3                                 & 68                    &                                & \multicolumn{1}{c|}{3}                     &                       & 4                                                                          & 41 \\
\textit{question} \#: &                       &                                & \multicolumn{1}{c|}{\textit{77, 89, 94}}   &                       & \textit{44, 65, 112, 113}                                                  &    \\ \hline
\multicolumn{7}{c}{\textbf{Agreement between observers}}                                                                                                                                                                                               \\ \cline{3-3} \cline{6-6}
                                       & \multicolumn{1}{c|}{} & \multicolumn{1}{c|}{Obs. 2}    & \multicolumn{1}{c|}{}                      & \multicolumn{1}{c|}{} & \multicolumn{1}{c|}{Obs. 2}                                                &    \\ \cline{3-4} \cline{6-7} 
                                       &                       & Y                              & \multicolumn{1}{c|}{N}                     &                       & Y                                                                          & N  \\ \cline{1-1}
\multicolumn{1}{|c|}{Obs. 1}           & Y                     & 64                             & \multicolumn{1}{c|}{2}                     &                       & 0                                                                          & 8  \\ \cline{1-1}
\multicolumn{1}{c|}{}                  & N                     & 4                              & \multicolumn{1}{c|}{1}                     &                       & 3                                                                          & 34 \\
                                       &                       & \multicolumn{2}{c}{AC1 = 0.9054 ($p << 0.001)$}                             &                       & \multicolumn{2}{c}{AC1 = 0.6888 ($p = 2.57 \times 10^{-8})$}                    \\ \cline{3-3} \cline{6-6}
                                       & \multicolumn{1}{c|}{} & \multicolumn{1}{c|}{Obs. 3}    & \multicolumn{1}{c|}{}                      & \multicolumn{1}{c|}{} & \multicolumn{1}{c|}{Obs. 3}                                                &    \\ \cline{3-4} \cline{6-7} 
                                       &                       & Y                              & \multicolumn{1}{c|}{N}                     &                       & Y                                                                          & N  \\ \cline{1-1}
\multicolumn{1}{|c|}{Obs. 1}           & Y                     & 65                             & \multicolumn{1}{c|}{1}                     &                       & 1                                                                          & 7  \\ \cline{1-1}
\multicolumn{1}{c|}{}                  & N                     & 3                              & \multicolumn{1}{c|}{2}                     &                       & 3                                                                          & 34 \\
                                       &                       & \multicolumn{2}{c}{AC1 = 0.9370 ($p << 0.001)$}                             &                       & \multicolumn{2}{c}{AC1 = 0.7110 ($p = 5.57 \times 10^{-9})$}                    \\ \cline{3-3} \cline{6-6}
                                       & \multicolumn{1}{c|}{} & \multicolumn{1}{c|}{Obs. 3}    & \multicolumn{1}{c|}{}                      & \multicolumn{1}{c|}{} & \multicolumn{1}{c|}{Obs. 3}                                                &    \\ \cline{3-4} \cline{6-7} 
                                       &                       & Y                              & \multicolumn{1}{c|}{N}                     &                       & Y                                                                          & N  \\ \cline{1-1}
\multicolumn{1}{|c|}{Obs. 2}           & Y                     & 66                             & \multicolumn{1}{c|}{2}                     &                       & 0                                                                          & 3  \\ \cline{1-1}
\multicolumn{1}{c|}{}                  & N                     & 2                              & \multicolumn{1}{c|}{1}                     &                       & 4                                                                          & 38 \\
                                       &                       & \multicolumn{2}{c}{AC1 = 0.9387 ($p << 0.001)$}                             &                       & \multicolumn{2}{c}{AC1 = 0.8184 ($p = 2.26 \times 10^{-14})$}                   \\ \hline
\multicolumn{7}{c}{\textbf{Incorrect interpretation or incoherent alternative choice?}}                                                                                                                                                                \\ \cline{3-3} \cline{6-6}
                                       & \multicolumn{1}{c|}{} & \multicolumn{1}{c|}{Coherence} & \multicolumn{1}{c|}{}                      & \multicolumn{1}{c|}{} & \multicolumn{1}{c|}{Coherence}                                             &    \\ \cline{3-4} \cline{6-7} 
                                       &                       & Y                              & \multicolumn{1}{c|}{N}                     &                       & Y                                                                          & N  \\ \cline{1-1}
\multicolumn{1}{|c|}{Interpretation}   & Y                     & 66                             & \multicolumn{1}{c|}{1}                     &                       & 16                                                                         & 2  \\ \cline{1-1}
\multicolumn{1}{c|}{}                  & N                     & 4                              & \multicolumn{1}{c|}{0}                     &                       & 25                                                                         & 2  \\
                                       &                       & \multicolumn{2}{c}{AC1 = 0.9244 ($p << 0.001)$}                             &                       & \multicolumn{2}{c}{AC1 = -0.0941 ($p = 0.5921)$}                                \\ \hline
\end{tabular}
}
\end{table}

\clearpage
\begin{table}[ht]
\centering
\caption{Summary statistics for Experiment 3: agreement between justificatives and potential harm to patients per observer (top part) and between observers (intermediate and bottom parts. Agreement by Gwet's AC1.}
\label{tab:agreement2}
\resizebox{\textwidth}{!}
{
\begin{tabular}{ccccccc}
\hline
\multicolumn{7}{c}{\textbf{Model correct}}                                                                                                                                                                  \\
                                    &                       & Y                           & \multicolumn{1}{c|}{}   &                       & N                                          &                  \\
                                    &                       & 71                          & \multicolumn{1}{c|}{}   &                       & 45                                         &                  \\ \hline
\multicolumn{7}{c}{\textbf{Intraobserver agreement (justificative x harm to patient)}}                                                                                                                      \\ \cline{3-3} \cline{6-6}
\textbf{Obs. 1}                     & \multicolumn{1}{c|}{} & \multicolumn{1}{c|}{Harm}   & \multicolumn{1}{c|}{}   & \multicolumn{1}{c|}{} & \multicolumn{1}{c|}{Harm}                  &                  \\ \cline{3-4} \cline{6-7} 
                                    &                       & Y                           & \multicolumn{1}{c|}{N}  &                       & Y                                          & N                \\ \cline{1-1}
\multicolumn{1}{|c|}{Justificative} & Y                     & 1                           & \multicolumn{1}{c|}{65} &                       & 7                                          & 1                \\ \cline{1-1}
\multicolumn{1}{c|}{}               & N                     & 4                           & \multicolumn{1}{c|}{1}  &                       & 35                                         & 2                \\
                                    &                       & \multicolumn{2}{l}{AC1 = -0.9437 ($p << 0.0001)$}     &                       & \multicolumn{2}{l}{AC1 = -0.5805 ($p = 6.98 \times 10^{-5})$} \\ \cline{3-3} \cline{6-6}
\textbf{Obs. 2}                     & \multicolumn{1}{c|}{} & \multicolumn{1}{c|}{Harm}   & \multicolumn{1}{c|}{}   & \multicolumn{1}{c|}{} & \multicolumn{1}{c|}{Harm}                  &                  \\ \cline{3-4} \cline{6-7} 
                                    &                       & Y                           & \multicolumn{1}{c|}{N}  &                       & Y                                          & N                \\ \cline{1-1}
\multicolumn{1}{|c|}{Justificative} & Y                     & 0                           & \multicolumn{1}{c|}{68} &                       & 3                                          & 0                \\ \cline{1-1}
\multicolumn{1}{c|}{}               & N                     & 0                           & \multicolumn{1}{c|}{3}  &                       & 32                                         & 10               \\
                                    &                       & \multicolumn{2}{l}{AC1 = -0.9121 ($p << 0.0001)$}     &                       & \multicolumn{2}{l}{AC1 = -0.3886 ($p = 0.0139)$}              \\ \cline{3-3} \cline{6-6}
\textbf{Obs. 3}                     & \multicolumn{1}{c|}{} & \multicolumn{1}{c|}{Harm}   & \multicolumn{1}{c|}{}   & \multicolumn{1}{c|}{} & \multicolumn{1}{c|}{Harm}                  &                  \\ \cline{3-4} \cline{6-7} 
                                    &                       & Y                           & \multicolumn{1}{c|}{N}  &                       & Y                                          & N                \\ \cline{1-1}
\multicolumn{1}{|c|}{Justificative} & Y                     & 0                           & \multicolumn{1}{c|}{68} &                       & 0                                          & 4                \\ \cline{1-1}
\multicolumn{1}{c|}{}               & N                     & 3                           & \multicolumn{1}{c|}{0}  &                       & 41                                         & 0                \\
                                    &                       & \multicolumn{2}{l}{AC1 = -1.0000 ($p << 0.0001)$}     &                       & \multicolumn{2}{l}{AC1 = -1.0000 ($p << 0.0001)$}             \\ \hline
\multicolumn{7}{c}{\textbf{Interobserver agreement (Harm to patient)}}                                                                                                                                      \\ \cline{3-3} \cline{6-6}
\textbf{}                           & \multicolumn{1}{c|}{} & \multicolumn{1}{c|}{Obs. 2} & \multicolumn{1}{c|}{}   & \multicolumn{1}{c|}{} & \multicolumn{1}{c|}{Obs. 2}                &                  \\ \cline{3-4} \cline{6-7} 
                                    &                       & Y                           & \multicolumn{1}{c|}{N}  &                       & Y                                          & N                \\ \cline{1-1}
\multicolumn{1}{|c|}{Obs. 1}        & Y                     & 0                           & \multicolumn{1}{c|}{5}  &                       & 34                                         & 8                \\ \cline{1-1}
\multicolumn{1}{c|}{}               & N                     & 0                           & \multicolumn{1}{c|}{66} &                       & 1                                          & 2                \\
                                    &                       & \multicolumn{2}{l}{AC1 =  0.9244 ($p << 0.0001)$}     &                       & \multicolumn{2}{l}{AC1 =  0.7343 ($p = 9.43\times 10^{-10})$}  \\ \cline{3-3} \cline{6-6}
\textbf{}                           & \multicolumn{1}{c|}{} & \multicolumn{1}{c|}{Obs. 3} & \multicolumn{1}{c|}{}   & \multicolumn{1}{c|}{} & \multicolumn{1}{c|}{Obs. 3}                &                  \\ \cline{3-4} \cline{6-7} 
                                    &                       & Y                           & \multicolumn{1}{c|}{N}  &                       & Y                                          & N                \\ \cline{1-1}
\multicolumn{1}{|c|}{Obs. 1}        & Y                     & 1                           & \multicolumn{1}{c|}{4}  &                       & 39                                         & 3                \\ \cline{1-1}
\multicolumn{1}{c|}{}               & N                     & 2                           & \multicolumn{1}{c|}{64} &                       & 2                                          & 1                \\
                                    &                       & \multicolumn{2}{l}{AC1 =  0.9054 ($p << 0.0001)$}     &                       & \multicolumn{2}{l}{AC1 =  0.8703 ($p << 0.0001)$}             \\ \cline{3-3} \cline{6-6}
\textbf{}                           & \multicolumn{1}{c|}{} & \multicolumn{1}{c|}{Obs. 3} & \multicolumn{1}{c|}{}   & \multicolumn{1}{c|}{} & \multicolumn{1}{c|}{Obs. 3}                &                  \\ \cline{3-4} \cline{6-7} 
                                    &                       & Y                           & \multicolumn{1}{c|}{N}  &                       & Y                                          & N                \\ \cline{1-1}
\multicolumn{1}{|c|}{Obs. 2}        & Y                     & 0                           & \multicolumn{1}{c|}{0}  &                       & 33                                         & 2                \\ \cline{1-1}
\multicolumn{1}{c|}{}               & N                     & 3                           & \multicolumn{1}{c|}{68} &                       & 8                                          & 2                \\
                                    &                       & \multicolumn{2}{l}{AC1 =  0.9559 ($p << 0.0001)$}     &                       & \multicolumn{2}{l}{AC1 =  0.6986 ($p = 1.96 \times 10^{-8})$}  \\ \hline
\end{tabular}
}
\end{table}

\clearpage
\section{Figures}

\end{document}